\documentclass{article}

\usepackage{microtype}
\usepackage{graphicx}
\usepackage{subfigure}
\usepackage{booktabs} 
\usepackage{amsmath}
\usepackage{amssymb}
\usepackage{bm}

\usepackage[ruled,vlined,linesnumbered]{algorithm2e}
\SetKwInput{Parameters}{Parameters}
\SetKwInput{Variables}{Variables}

\newcommand{\vect}[1]{\mathbf{#1}}

\newcommand{\br}[1]{\mathopen{}\left(#1\right)\mathclose{}}
\newcommand{\set}[1]{\left\{#1\right\}}

\newcommand{\prob}[1]{p\br{#1}}

\newcommand{\g}{\,|\,}

\usepackage{hyperref}

\def\model{\mathcal{M}}

\usepackage[accepted]{icml2021}

\icmltitlerunning{Optimal simulation-based Bayesian decisions}

\begin{document}

\twocolumn[
\icmltitle{Optimal simulation-based Bayesian decisions}



\icmlsetsymbol{equal}{*}

\begin{icmlauthorlist}
\icmlauthor{Justin Alsing}{calda,loo}
\icmlauthor{Thomas D. P. Edwards}{calda,to}
\icmlauthor{Benjamin Wandelt}{calda,goo,four}
\end{icmlauthorlist}

\icmlaffiliation{calda}{Calda AI AB, Villa Bellona, Universitetsv\"{a}gen 8, 114 18 Stockholm, Sweden}
\icmlaffiliation{loo}{Oskar Klein Centre for Cosmoparticle Physics, Stockholm
University, Stockholm SE-106 91, Sweden}
\icmlaffiliation{to}{William H. Miller III Department of Physics and Astronomy, Johns Hopkins University, Baltimore, Maryland 21218, USA}
\icmlaffiliation{four}{Center for Computational Astrophysics, Flatiron Institute, 162 5th Avenue, New York, NY 10010, USA}
\icmlaffiliation{goo}{Sorbonne Universit\'e, CNRS, UMR 7095, Institut d’Astrophysique de Paris, 98 bis boulevard Arago, 75014 Paris, France}

\icmlcorrespondingauthor{Justin Alsing}{justin.alsing@fysik.su.se}

\icmlkeywords{Machine Learning, ICML}

\vskip 0.3in
]



\printAffiliationsAndNotice{}  

\begin{abstract}
We present a framework for the efficient computation of optimal Bayesian decisions under intractable likelihoods, by learning a surrogate model for the expected utility (or its distribution) as a function of the action and data spaces. We leverage recent advances in simulation-based inference and Bayesian optimization to develop active learning schemes to choose where in parameter and action spaces to simulate. This allows us to learn the optimal action in as few simulations as possible. The resulting framework is extremely simulation efficient, typically requiring fewer model calls than the associated posterior inference task alone, and a factor of $100-1000$ more efficient than Monte-Carlo based methods. Our framework opens up new capabilities for performing Bayesian decision making, particularly in the previously challenging regime where likelihoods are intractable, and simulations expensive.
\end{abstract}

\section{Introduction}
Making optimal decisions in the face of uncertainty is the end-game of probabilistic reasoning from data. Bayesian decision theory provides the unique framework for making rational decisions (under a reasonable set of axioms, \citealp{von2007theory}), and is postulated to be fundamental to how human decision-making works in the brain \cite{kording2004bayesian, doya2007bayesian, doya2008modulators, funamizu2016neural, lindig2022bayes}.

In the Bayesian paradigm, optimal decision making involves finding the action, $\bm{a}^*$, that maximizes the expectation value of some utility function $U(\bm{y}, \bm{a})$, with respect to the posterior predictive distribution $P(\bm{y} | \bm{x}, \bm{a}, \model)$ where $\bm{y}$ represents future outcomes corresponding to actions $\bm{a}$, $\bm{x}$ is our observations, and $\model$ represents our model assumptions (see eg., \citealp{berger2013statistical} for a review). Hence, the goal is to find the action that maximizes the following integral:
\begin{eqnarray}
\label{eq:expected_utility}
    \bm{a}^* &=& \mathrm{argmax}_a ~ u(\bm{a}, \bm{x},  \model)\,, \nonumber \\
    &=&\mathrm{argmax}_a \int U(\bm{y}, \bm{a}) P(\bm{y} | \bm{a}, \bm{x}, \model) d\bm{y}\,, \nonumber \\
    &=& \mathrm{argmax}_a \int U(\bm{y}, \bm{a}) P(\bm{y} | \bm{x}, \bm{\theta}, \bm{a}, \model) P(\bm{x} | \bm{\theta}, \model) \nonumber \\
    &\,&\phantom{U(y, a) P(\bm{y} | \bm{\theta}) }\times P(\bm{\theta})d\bm{\theta} d\bm{y}\,,
\end{eqnarray}
where $\bm{\theta}$ are the model parameters describing the state of the world under model $\model$, and $P(\bm{\theta} | \bm{x}, \model)$ is the posterior distribution for the model parameters given the observations. For brevity, we will drop the implicit $\model$ in the notation. 

Finding the optimal action therefore involves optimizing a typically high-dimensional integral (over both the data and parameter spaces). Na\"{i}vely, this requires three individually challenging computational steps: (1) \emph{inference}, i.e., sampling or approximating the posterior $P(\bm{\theta} | \bm{x})$, (2) \emph{integration}, i.e., estimating the integral for the posterior predictive expected utility $u(\bm{a}, \bm{x})$ in Eq.~\eqref{eq:expected_utility}, and (3) \emph{optimization}, i.e., maximize $u(\bm{a}, \bm{x})$ with respect to the action $\bm{a}$. This multi-step process presents a substantial computational challenge. Nonetheless, when the likelihood function is tractable, Bayesian methods for sampling or approximating the posterior provide a path forward for approximating the required integrals by Monte Carlo (or otherwise), and performing the optimization task, albeit at significant computational cost. In spite of the central importance of Bayesian decision theory, methods for computing optimal actions in the Bayesian setting have been mostly limited to inefficient Monte Carlo based approaches, and there is limited literature on developing improved computational strategies.

To make matters more challenging still, we are often in a situation where we have a forward model which can be easily simulated, but does not have a tractable likelihood. In these cases the model is specified \textit{implicitly} by a generative model, such as computer simulations or a neural generator. It may also be implicit in a set of labeled parameter-data pairs.  In the literature this setting is often referred to as \textit{likelihood-free} or \textit{simulation-based}. In these cases the posterior itself is intractable and needs to be obtained through implicit inference techniques, leading to an additional level of difficulty in integrating and optimizing $u(\bm{a}, \bm{x})$.

One approach to performing implicit inference is to learn the sampling distribution of the observations $P(\bm{x} | \bm{\theta})$ or the posterior $P(\bm{\theta} | \bm{x})$ from simulated parameter-data draws $\{\bm{\theta}, \bm{x}\}$ of the forward model \cite{papamakarios2016fast, alsing2018massive, papamakarios2019sequential, lueckmann2019likelihood, alsing2019fast}. Phrasing posterior inference as a conditional density-estimation task in this way has led to dramatic improvements in likelihood-free inference compared to previous approximate Bayesian computation (ABC) based approaches (see \citealp{beaumont2019approximate} for a review). These improvements are largely driven by recent advances in neural network based conditional density estimators (such as Normalizing Flows; \citealp{papamakarios2021normalizing, 2016arXiv160508803D, 2019arXiv190604032D, 2015arXiv150203509G, 2017arXiv170507057P, 2018arXiv180703039K,2018arXiv181001367G}).

\begin{figure*}[ht!]
    \centering
    \includegraphics[width=15cm]{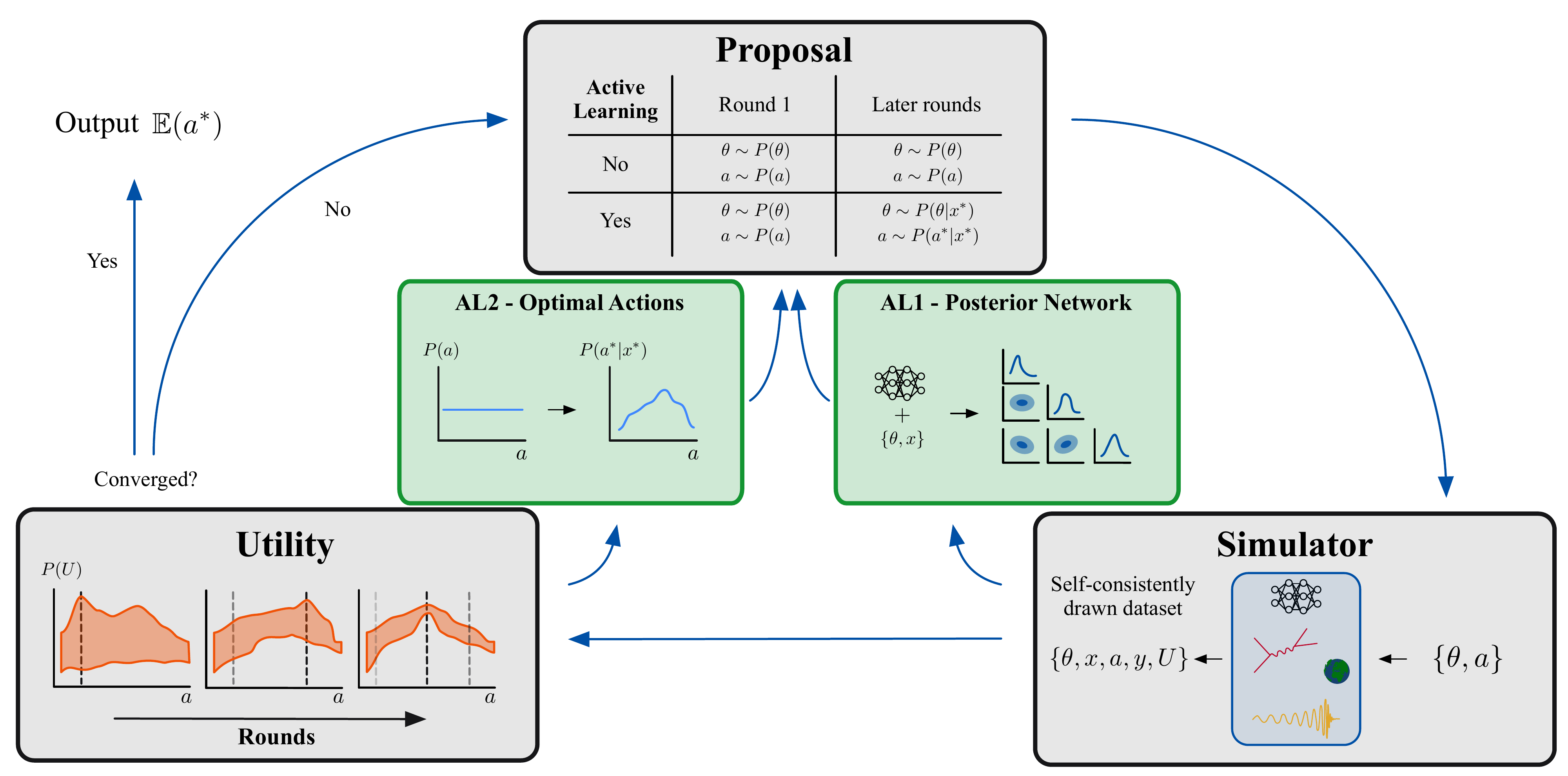}
    \caption{Overview of the various parts of the learning algorithm discussed in \S \ref{sec:regression}-\ref{sec:cde}. Grey boxes represent the three core parts of the algorithm with blue arrows showing the flow of information. Green boxes represent the two active learning strategies designed to reduce the simulation burden of the overall algorithm (see \S\,\ref{sec:active} for further discussion).}
    \label{fig:graphic_overview}
\end{figure*}

In this paper, we leverage these recent advances in simulation-based inference (SBI) to introduce a new framework for making optimal Bayesian decisions: hereafter, we will refer to this framework as SBD (simulation-based decision-making). Importantly, we bypass the need to perform posterior inference and compute the posterior predictive expected utility integrals directly. By relying only on forward simulations, our framework can handle both the tractable-likelihood and likelihood-free regimes whilst remaining extremely simulation efficient. In fact, we will show that finding the optimal action often requires fewer model calls than would be required to perform the associated posterior inference step alone. This enables new capabilities for making optimal Bayesian decisions in situations where forward simulation is expensive.

We introduce two approaches to efficiently find $\bm{a}^*$. Firstly, we show that a surrogate model for $u(\bm{a}, \bm{x})$ can be learnt via regression. The optimal action can then be simply found by optimizing the surrogate model (evaluated at the observed data $\bm{x}^*$). Alternatively, we cast the problem as a one-dimensional conditional density-estimation task, learning the posterior predictive distribution of the utility as a function of action and observations (borrowing ideas from density-estimation likelihood-free inference); this is particularly useful when the utility distribution is significantly non-Gaussian and heterogeneous, which would otherwise make for a challenging regression task.

When parameter-data pairs can be generated on demand,  both these approaches can use active learning to optimize where in parameter-space and action-space to perform simulations. In order to learn the expected utility in the neighbourhood of the observed data with as few simulations as possible, we exploit density-estimation likelihood-free inference to learn an interim approximation for the posterior, which acts as an adaptive proposal for drawing simulation parameters. For optimizing which actions to simulate, we leverage Bayesian inference to obtain a posterior distribution for the expected utility surface given the available training data at any given stage; this in turn provides us with a posterior distribution for the optimal action, which can be used as an adaptive proposal for drawing which actions to simulate next. We show that these active learning schemes result in a dramatic reduction in the number of simulation calls required to make optimal Bayesian decisions.

Reinforcement learning (RL) has revolutionized the way we perform sequential decision tasks (see \citealp{sutton2018reinforcement} for a review). In RL, one is concerned with constructing an optimal decision policy which is amortized with respect to the observations it is conditioned on, i.e., once trained it enables rapid decision making in a sequential setting. However, the exploration phase in RL typically requires the ability to generate large amounts of training data on-the-fly. In this paper we are concerned with computing optimal decisions in a different setting: making isolated (static) decisions, based on limited observations (i.e., where the posterior uncertainties over the model and its parameters are substantial), in the regime where likelihoods may be intractable and simulations expensive.

The paper is structured as follows: we show how Bayesian decision making can be cast as a regression problem in \S\,\ref{sec:regression}, or a conditional density-estimation task in \S\,\ref{sec:cde}. In \S\,\ref{sec:active} we show how active learning can be employed to optimize where in parameter space to run simulations and which actions to explore in order to reduce the number of model calls required to find $\bm{a}^*$. In \S\,\ref{sec:wh}-\ref{sec:lv} we demonstrate the framework on two examples: planning a stock purchase in retail, and managing deer populations in ecology. Figure~\ref{fig:graphic_overview} provides a high level overview of the overall framework. We will refer to various parts of this figure as they are introduced throughout the text.

\textit{Glossary:} This paper involves a number of definitions and moving parts. To assist the reader we compile some of the terms used and our notation for easy referral.
\begin{itemize}
    \itemsep 0pt
    \item \textit{Conditional Density Estimation:} Here we are estimating the density for some random variable $\bm{x}$, conditioned on some other variable $\bm{\theta}$ i.e., $p(\bm{x}|\bm{\theta})$. Since this may or may not involve a neural network, conditional density estimation also encompasses neural density estimation.  This is typically used to bypass the need to write down a tractable likelihood i.e., to perform implicit or likelihood-free inference.
    \item \textit{Posterior predictive expected utility:} We use this term to denote $u(\bm{a}, \bm{x})$ defined in Eq.~\eqref{eq:expected_utility}. It is the expectation value of the utility function $U(\bm{y}, \bm{a})$, taken with respect to the posterior predictive distribution $P(\bm{y} | \bm{x}, \bm{a})$ for future outcomes $\bm{y}$ given past observations $\bm{x}$ and action $\bm{a}$.
    \item \textit{Simulator:} Throughout this work, the simulator is treated as a black box that takes inputs $\{\bm{\theta}, \bm{a}\}$ and outputs simulated observations $\bm{x}$, future outcomes $\bm{y}$, and the associated utility $U(\bm{y}, \bm{a})$.
    \item \textit{Parameter/Action Spaces:} We use parameter-space to refer the set of parameters $\bm{\theta}$ used by our simulator (which are unknown and need inferring from observations). 
    The action space refers to the space of possible actions. 
\end{itemize}

\section{Bayesian decision making as a regression task}
\label{sec:regression}
Following Eq.~\eqref{eq:expected_utility}, making optimal decisions involves finding the action $\bm{a}$ that maximizes the posterier predictive expectated utility, $u(\bm{a}, \bm{x})$. Since the integral in Eq.~\eqref{eq:expected_utility} is typically high-dimensional and cannot be computed directly, we instead seek a means to learn a surrogate model $\hat{u}(\bm{a}, \bm{x})$ for the expected utility \emph{without} having to compute any integrals. We can estimate $\hat{u}(\bm{a}, \bm{x})$ by regressing it against realizations of the utility $U(\bm{y}, \bm{a})$ from the training data $\bm{y}\sim P(\bm{y} | \bm{x}, \bm{a})$, using a mean square loss, as demonstrated below. 

Consider the square error of a function $\hat{u}(\bm{a}, \bm{x})$ with respect to realizations of the utility $U(\bm{y}, \bm{a})$, under the posterior predictive distribution for future outcomes $P(\bm{y} | \bm{x}, \bm{a})$:
\begin{eqnarray}
    \mathbb{E}_{P(\bm{y} | \bm{x}, \bm{a})}\left[(U(\bm{y}, \bm{a}) - \hat{u}(\bm{a}, \bm{x}))^2\right] \equiv \mathcal{L}(\bm{y}, \bm{a}, \bm{x})\,.
\end{eqnarray}
Minimizing the square error with respect to $\hat{u}(\bm{a}, \bm{x})$ gives:
\begin{eqnarray}
\label{eq:mse_proof}
    \underset{\hat{u}(\bm{a}, \bm{x})}{\mathrm{argmin}} \left\{\mathcal{L}(\bm{y}, \bm{a}, \bm{x})\right\}\equiv u(\bm{a}, \bm{x})\,.
\end{eqnarray}

Therefore, the function which minimizes the square error with respect to $U(\bm{y}, \bm{a})$ gives an unbiased estimator for the expected utility, $u(\bm{a}, \bm{x})$\footnote{Assuming the usual caveats of a sufficiently expressive model/network, and sufficient training data and training.}.

In practice, the expected utility can hence be learned as follows. Generate training data by drawing parameters from the prior $\bm{\theta} \sim P(\bm{\theta})$, simulating observations from their (implicit) sampling distribution $\bm{x} \sim P(\bm{x} | \bm{\theta})$, choosing actions $\bm{a}$ from some prior distribution $P(\bm{a})$, simulating future outcomes from their (implicit) sampling distribution $\bm{y}\sim P(\bm{y} | \bm{\theta}, \bm{a})$, and finally computing the utility for each future outcome-action pair, $U(\bm{a}, \bm{y} )$. The resulting set of $\{\bm{y, a, x, \theta}, U\}$ are implicitly drawn from the joint density $P(\bm{x}, \bm{\theta}, \bm{y} | \bm{a})$ (and $U$ is deterministic given $\bm{y}$ and $\bm{a}$). In Fig.~\ref{fig:graphic_overview}, this sampling process corresponds to the top and bottom right grey boxes. A parametric model (such as a neural network) for the expected utility $\hat{u}(\bm{x}, \bm{a};\bm{\psi})$ can then be fit by minimizing the square error,
\begin{eqnarray}
    \mathcal{L}_\mathrm{MSE} = \sum_{i=1}^N \left[\hat{u}(\bm{x}_i, \bm{a}_i;\bm{\psi}) - U(\bm{y}_i, \bm{a}_i)\right]^2\,,
\end{eqnarray}
with respect to the model parameters $\bm{\psi}$.

In the non-parametric limit with Gaussian priors on the utility surface, the expected utility can instead be modeled as a Gaussian process (GP; see \citealp{rasmussen2006gaussian} for a review). GPs bring the convenience of an analytically tractable posterior distribution for the utility surface, given the training data.

The GP posterior for the expected utility at points $\mathbf{z}'=(\mathbf{a}', \mathbf{x}')$, given training data $\{\mathbf{u}, \mathbf{z}=(\mathbf{a}, \mathbf{x})\}$, is Gaussian with mean:
\begin{eqnarray}
\label{gp-mean}
\mu(\mathbf{z}') + K(\mathbf{z}', \mathbf{z})\left[K(\mathbf{z}, \mathbf{z}) + \sigma^2\mathbf{I}\right]^{-1}(\mathbf{u} - \mu(\mathbf{z}'))\,,
\end{eqnarray}
and covariance:
\begin{eqnarray}
\label{gp-covariance}
K(\mathbf{z}', \mathbf{z}') - K(\mathbf{z}', \mathbf{z})\left[K(\mathbf{z}, \mathbf{z}) + \sigma^2\mathbf{I}\right]^{-1}K(\mathbf{z}, \mathbf{z}')\,,
\end{eqnarray}
where $\mu(\mathbf{z})$ is the assumed mean of the GP prior, $\sigma$ is a hyper-parameter characterizing the variance of the utility distribution, and $K$ is the kernel\footnote{Throughout we will use a squared exponential kernel, defined as $K(\bm{x}_i, \bm{x}_j) = A e^{-\frac{1}{2}\left[(\bm{x}_i - \bm{x}_j)\odot\bm{\ell}^{-1}\right]^2}$, where the amplitude $A$ and the length-scales $\bm{\ell}$ (per input dimension) are the kernel hyper-parameters, and $\odot$ denotes element-wise multiplication.} defining the covariance of the GP prior (which in turn is described by hyper-parameters $\eta$). All hyper-parameters $\lambda = (\eta, \sigma)$ are typically fit by maximizing the likelihood:
\begin{eqnarray}
\label{gp-loglikelihood}
P(\mathbf{u} | \mathbf{z}, \lambda) &=& \frac{1}{\sqrt{|2\pi C(\mathbf{z}, \mathbf{z})|}}\nonumber \\
&& \times e^{-\frac{1}{2}(\mathbf{u}-\mu(\mathbf{z}))^\mathrm{T}C(\mathbf{z}, \mathbf{z})^{-1}(\mathbf{u}-\mu(\mathbf{z}))}
\end{eqnarray}
where $C(\mathbf{z}, \mathbf{z}) = K(\mathbf{z}, \mathbf{z}) + \sigma^2\mathbf{I}$.

Once a model for the expected utility is fit (either parametric or non-parametric), it can be evaluated at the observations $\bm{x}^*$, and maximized with respect to action $\bm{a}$ to find the optimal action. By casting the decision problem as a regression task in this fashion, we neither have to perform posterior inference or approximate the integral for the expected utility explicitly; both are taken care of implicitly.
\section{Bayesian decision making as a conditional density estimation task}
\label{sec:cde}
As discussed in the previous section, minimizing the mean square error of $\hat{u}(\bm{a}, \bm{x})$ against the utility $U(\bm{y}, \bm{a})$, calculated with samples drawn from $y \sim P(\bm{y} | \bm{x}, \bm{a})$, provides an unbiased estimator for $u(\bm{a}, \bm{x})$ for a given combination of data $\bm{x}$ and action $\bm{a}$ (c.f Equation \ref{eq:mse_proof}). However, if the distribution of the utility $U$ over the action and observable spaces is non-Gaussian and heterogeneous, then performing a least-square model fit for the expected utility across the action and data space may result in biased, sub-optimal or unstable training\footnote{Particularly in the case of limited training data.} (due to, for example, the outsized effect of tail events in the training data).

In the context of learning the posterior predictive expected utility for decision making, we can resolve this problem by instead learning the distribution of the utility conditioned on the action and observations, $P(U | \bm{a}, \bm{x})$, in such a way that a surrogate model for the expectation $\hat{u}(\bm{a}, \bm{x})$ is obtained explicitly. In this way, Bayesian decision making is cast as a conditional density estimation task, in the same spirit as density-estimation based approaches to likelihood-free inference \cite{papamakarios2016fast, alsing2018massive, papamakarios2019sequential, lueckmann2019likelihood, alsing2019fast}. Note that the idea of learning the utility distribution, rather that just regressing its expectation, is already in use in a similar context in distributional reinforcement learning \cite{bellemare2017distributional}.

We generate training data in the same way as in the regression approach, resulting in a set of data $\{\bm{y}, \bm{a}, \bm{x}, \bm{\theta}, U\}$ which is self-consistently drawn from $P(\bm{x}, \bm{\theta}, \bm{y} | \bm{a})$ (c.f. \S \ref{sec:regression}). The utility distribution can then parameterized as a neural conditional density estimator $\hat{P}(U | \bm{a}, \bm{x})$ (such as a mixture density network or normalizing flow), which is trained by maximum likelihood, i.e., by minimizing the negative log-likelihood:
\begin{eqnarray}
    \mathcal{L}_\mathrm{ML} = -\sum_{i=1}^N \mathrm{ln}\, \hat{P}(U_i | \bm{a}_i, \bm{x}_i)\,.
\end{eqnarray}
As with the regression approach, phrasing the decision problem in this way bypasses the need to perform inference and compute integrals for the expected utility explicitly. The decision task is reduced to learning a one-dimensional density conditioned on the action and data spaces, whereas the inference step alone (in the likelihood-free setting) typically involves estimating a $\mathrm{dim}(\bm{\theta})$ dimensional density conditional on $\mathrm{dim}(\bm{x})$ data. This means that (provided the dimensionality of the action space is not too large) the complexity of the learning task for decision-making is typically lower than for the associated inference problem alone. Furthermore, in order to find the correct optimal action, the utility distribution does not need to be determined with high fidelity, nor does the density estimate need to be unbiased; all we require is that we can correctly locate the maximum of the expected utility in the space of actions.

As for simulation-based inference, it is generally desirable to compress the data-vector $\bm{x}$ down to as small a number of informative summary statistics as possible, to reduce the dimensionality of the learning task. In practise, this typically means that observations are compressed to $\mathrm{dim}(\bm{\theta})$ summaries -- one per parameter of interest -- while preserving as much information about the parameters of interest as possible \cite{alsing2018generalized,charnock2018automatic, alsing2019nuisance,jeffrey2020solving, hoffmann2022minimizing}. When performing simulation-based decision-making we are able to exploit these same methods from the simulation-based inference literature for compressing large data vectors down to a manageable number of summaries. We note that in the decision-making problem, we are primarily interested in the utility of future outcomes rather than the latent model parameters describing our system; therefore, there may exist even better compression schemes that aim to preserve only the information content regarding future outcomes (rather than all of the latent model parameters). We leave this to future work.
\section{Reducing the simulation burden with active learning}
\label{sec:active}
The procedures described in \S \ref{sec:regression}-\ref{sec:cde} would learn the expected utility amortized over the prior-predictive distribution for the observations $x\sim P(\bm{x})$, and over the full support of the action space sampled from when generating training data. 
In the case where the goal is to find the optimal action given some fixed observations $\bm{x}^*$,  we can reduce the overall simulation cost by focusing on learning the expected utility in the neighbourhood of $\bm{x}^*$ and $\bm{a}^*$ only. 
To this end, we introduce two active learning schemes -- one to optimize where in parameter space to run simulations (so that we focus on the neighborhood of $\bm{x}^*$), and one to optimize where in action space to simulate future outcomes (to focus on the neighborhood of $\bm{a}^*$). These two active learning schemes are illustrated by the green boxes in Fig.~\ref{fig:graphic_overview}.

\subsection{Interim likelihood-free posterior inference for optimizing where to run simulations}
Rather than drawing parameters from the prior, we can instead draw parameters from an adaptive proposal density $q(\bm{\theta})$ that represents our current approximation of the posterior $P(\bm{\theta} | \bm{x}^*)$, ensuring that simulated observations preferentially occupy the neighbourhood of $\bm{x}^*$ that we are most interested in (in the same spirit as population Monte Carlo and sequential Monte Carlo methods, \citealp{DelMoral2006, Sisson2007, Toni2009, ToniStumpf2009, Beaumont2009, Bonassi2015}).

The adaptive parameter proposal scheme works as follows: we start proposing parameters from the prior $q_0(\bm{\theta}) = P(\bm{\theta})$ and generating $K$ draws of $\{\bm{\theta, x, y, a}, U\}$ (by simulation) as described above in \S\,\ref{sec:regression}. Then, we construct a neural density estimator for the (amortized) posterior $\hat{P}(\bm{\theta} | \bm{x})$, which is trained by maximum likelihood:
\begin{eqnarray}
    \mathcal{L} = -\sum_{i=1}^N \mathrm{ln}\, \hat{P}(\bm{\theta}_i | \bm{x}_i) w_i(\bm{\theta}_i),
\end{eqnarray}
where the importance weights $w(\bm{\theta}) = P(\bm{\theta}) / Q(\bm{\theta})$ are given by the ratio of the prior and the (global) proposal $Q$ from which parameters are drawn in the aggregated training set so far. Note that at the end of the first round, the importance weights are equal to unity (since the initial proposal is equal to the prior).

In the next round, we draw parameters from the updated proposal $q_1(\bm{\theta}) = \hat{P}(\bm{\theta} | \bm{x}^*)$, run new simulations, and retrain the amortized posterior on all of the simulations drawn up to that point. 
After $n$ rounds, the global proposal $Q(\bm{\theta})$ from which parameters are drawn in the aggregated training set is an equally-weighted mixture of $\{q_0(\bm{\theta}), \dots, q_n(\bm{\theta})\}$. Proceeding in this fashion, the amortized posterior becomes an increasingly accurate representation of the true posterior, and simulations are preferentially run in the most relevant region of parameter space.

As a result of drawing parameters from an adaptive proposal distribution rather than the prior, we must include importance weights to correct the loss functions for fitting the expected utility. The square error and maximum-likelihood losses for the regression and density-estimation approaches become (respectively):
\begin{eqnarray}
    &\mathcal{L}_\mathrm{MSE}& = -\sum_{i=1}^N \left[\hat{u}(\bm{x}_i, \bm{a}_i) - U(\bm{y}_i, \bm{a}_i)\right]^2 w(\bm{\theta}_i)\,,\nonumber \\ &\mathcal{L}_\mathrm{ML}& = -\sum_{i=1}^N \mathrm{ln}\, \hat{P}(U_i | \bm{a}_i, \bm{x}_i)\, w(\bm{\theta}_i)\,.
\end{eqnarray}
In the non-parametric (Gaussian process) limit, the importance weights can be included in Eqs.~(\ref{gp-mean}-\ref{gp-loglikelihood}) by replacing $\sigma^2\mathbf{I}\rightarrow \sigma^2\mathrm{diag}(\mathbf{w}^{-1})$~\cite{Wen2016WeightedGP}.
\subsection{Expected utility inference for optimizing which actions to simulate}
The second design choice in the context of active learning is how to choose which actions to simulate, given that we are only interested in learning the expected utility in the neighbourhood of the optimal action $\bm{a}^*$.

If instead of obtaining a point estimate for the expected utility surface we perform posterior inference with respect to it, then after any number of simulations we have a posterior distribution for the optimal action $\bm{a}^*$. This posterior then provides a natural adaptive proposal density for drawing new actions to simulate, encoding our current beliefs about the optimal action given the information we have available up to that point. This is in the spirit of Bayesian optimization \cite{shahriari2015taking, gutmann2016bayesian}.

If the expected utility (or utility distribution) is parameterized by a neural network, one would typically resort to approximate inference schemes to provide uncertainties over the predictions, such as variational inference \cite{blei2017variational}, dropout \cite{srivastava2014dropout}, or training ensembles of networks \cite{hansen1990neural}. Where the expected utility is learned as a GP, posterior uncertainties on the utility surface (under Gaussian assumptions) are available analytically [see Eq.~\eqref{gp-covariance}].

Combining the active learning strategies with the utility regression and density estimation approaches described in \S \ref{sec:regression} and \ref{sec:cde} is summarized in Algorithm \ref{alg:algotirhm1} and Figure \ref{fig:graphic_overview}. The framework of learning a surrogate model for the posterior expected utility over the action- and data-spaces via either regression or density estimation, with or without the active learning schemes described in \S \ref{sec:active}, we will refer to as SBD: simulation-based decision-making.

In the following sections we demonstrate SBD on two example problems: the warehousing problem in \S \ref{sec:wh}, and the Lotka-Volterra problem in \S \ref{sec:lv}.
\begin{algorithm}[t!]
initialize parameter proposal $Q(\bm{\theta})=P(\bm{\theta})$ \\
initialize action proposal $q_a(\bm{a}) = P(\bm{a})$ \\
\For{$n=1:N$}{
\For{$k=1:K$}{
sample $\bm{\theta}_k \sim Q\br{\bm{\theta}}$ \\
simulate $\vect{x}_k \sim \prob{\vect{x}\g \bm{\theta}_k}$
}
train amortized neural posterior $q_n\br{\bm{\theta}\g\vect{x}}$ on $\set{\bm{\theta}_n, \vect{x}_n}$\\
update $Q(\bm\theta) \rightarrow \mathrm{Mixture}(q_0\br{\bm{\theta}\g\vect{x}^*},\dots,q_n\br{\bm{\theta}\g\vect{x}^*})$ \\
\For{$k=1:K$}{
sample $\bm{a}_k \sim q_a\br{\bm{a}}$ \\
simulate $\vect{y}_k, U(\vect{y}_k, a) \sim \prob{\vect{k}\g \bm{\theta}_k, \bm{x}_k, \bm{a}_k}$
}
infer utility surface $P(\hat{u}(\bm{a}, \bm{x}) | \left\{u, \bm{a}, \bm{x}\right\})$ \\
update $q_a(\bm{a}) \rightarrow P(\bm{a}^* | \bm{x}^*)$ given $P(\hat{u}| \left\{u, \bm{a}, \bm{x}\right\})$
}
\caption{Learning the optimal action via inference of the expected utility surface, with an adaptive proposal for drawing parameters based on an amortized neural posterior estimator, and Bayesian optimization for choosing which actions to simulate.}
\label{alg:algotirhm1}
\end{algorithm}
\begin{table}[]
    \centering
    \begin{tabular}{c|l|l}
       \textbf{Problem} & \textbf{Method}  & \textbf{\# sims}
        \\
       \hline
        Warehousing & Posterior inference (SBI) & $\sim 100$  \\
        Warehousing & Monte-Carlo & $1.7\cdot10^{4}$  \\
        Warehousing & SBD, no active learning & $80$  \\
        Warehousing & \textbf{SBD, active learning} & $\mathbf{48}$  \\
        \hline
        Lotka-Volterra & Posterior inference (SBI) & $\sim 2000$  \\
        Lotka-Volterra & Monte-Carlo & $1.9\cdot10^{5}$  \\
        Lotka-Volterra & SBD, no active learning & $4096$  \\
        Lotka-Volterra & \textbf{SBD, active learning} & $\mathbf{1024}$   \\
    \end{tabular}
    \caption{Comparison of the total number of simulations required to compute the optimal Bayesian decision for the example problems considered in this work: the Warehousing problem \S \ref{sec:wh}, and the Lotka-Volterra problem \S \ref{sec:lv}. Monte-Carlo refers to performing optimization of a Monte-Carlo estimate for the posterior expected utility integral; in these cases, we choose the number of draws for evaluating the Monte-Carlo integral such that the action converges to the desired accuracy (1\% and 10\% for the Warehousing and Lotka-Volterra problems respectively). We also show the number of simulations required for the posterior inference step alone (using sequential amortized neural posterior estimation), for reference.}
    \label{tab:results}
\end{table}
\section{Example I: The warehousing problem}
\label{sec:wh}
\begin{figure*}[t!]
    \centering
    \includegraphics[width=16cm]{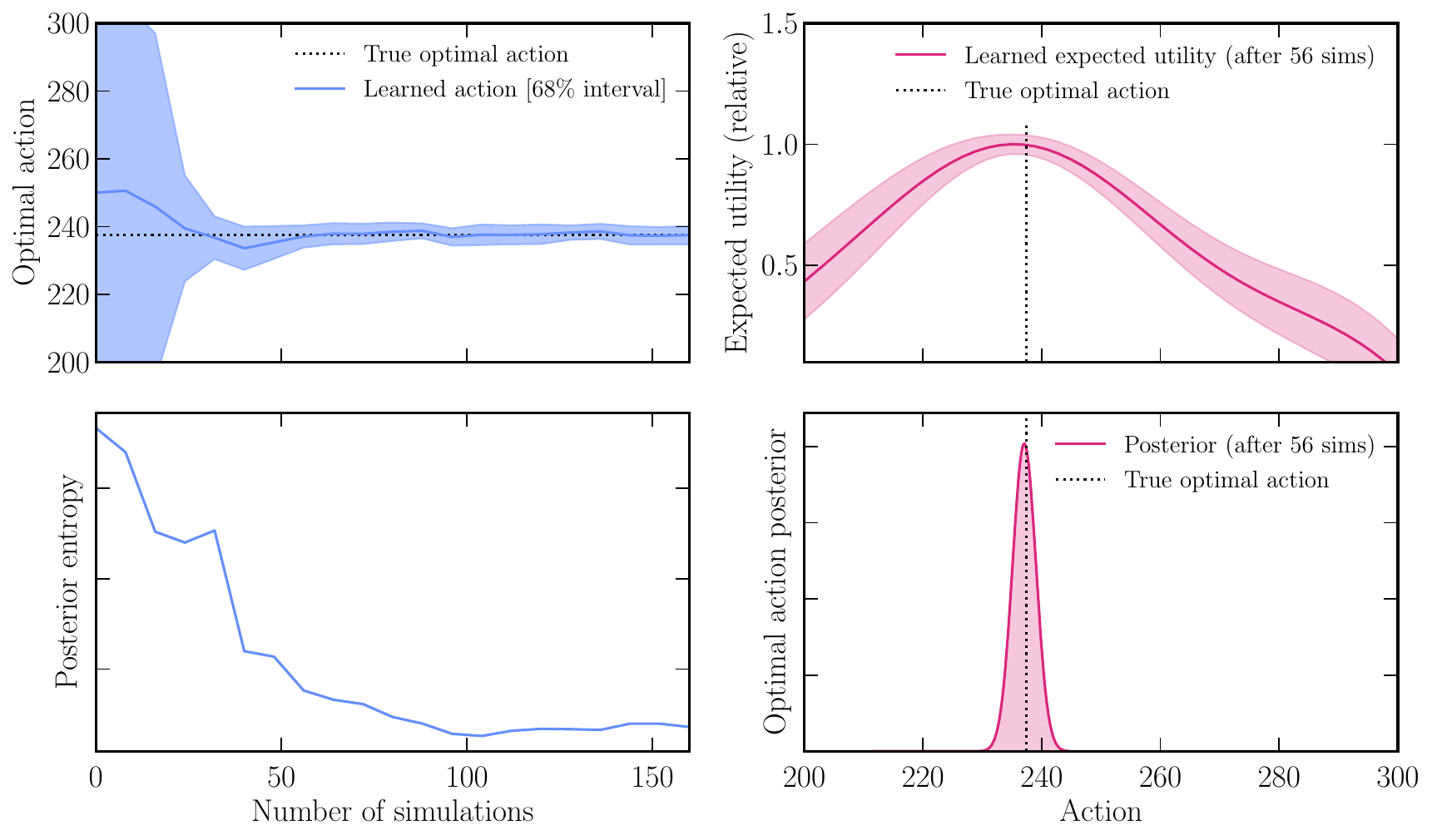}
    \caption{Top left: learned optimal action (with 68\% uncertainties) as a function of the number of simulations. Top right: converged learned expected utility surface (with 68\% uncertainties) as a function of action after $56$ simulations. Bottom right: the optimal action posterior after $56$ simulations. Bottom left: convergence of the amortized posterior (entropy) as a function of the number of simulations.}
    \label{fig:wh_results}
\end{figure*}
Suppose you are a retailer faced with deciding how much of a certain product to stock for the coming month, given an uncertain forecast for demand for the product. The goal is to maximize your overall profit -- i.e., sales, minus the costs of buying and storing the product --  while mitigating any additional negative effects of going out of stock (\textit{e.g.}, contractual out-of-stock penalties, or negative effects on customer loyalty). Suppose that you have data for the demand for the product over the previous twelve months, which you can use (with your assumed model for the demand time-series) to predict the demand for the coming month.

The \emph{action} $\bm{a}$ in this case is the amount of product you choose to stock, the \emph{outcome} $\bm{y}$ is the demand for the product for the coming month, which you can predict based on \emph{observations} $\bm{x}$ for the historical demand and your \emph{model} $\mathcal{M}$. For this simple case we will take as our model that the monthly demand is an uncorrelated Gaussian random variate, with unknown \emph{parameters} (mean and variance) $\bm{\theta}=(\mu, \sigma)$.

We take our \emph{utility} to be the net profit, minus an out-of-stock penalty, defined as:
\begin{eqnarray}
U(\bm{y}, \bm{a}) = V \mathrm{min}(\bm{a}, \bm{y}) - C\,\bm{a} - P\mathrm{max}(0, \bm{y}-\bm{a})\,.
\end{eqnarray}
where the wholesale cost (including storage) of each item is $C$, and the resale value of each item by $V$. In the eventuality that the demand exceeds the stock, we assume a penalty $P$ per item for being out-of-stock.

\subsection{Experiments}

The twelve months of historical demand data $\bm{x}$ are simulated assuming $\mu=234$, $\sigma=5$, and for the utility function we take $C=90$, $V=100$, and $P=100$. We assume priors $P(\mu) = N(230, 10)$ and $P(\sigma) = U(1, 10)$ on the model parameters, and $P(a) = U(200, 300)$ for the action.

We run the SBD algorithm described in Figure \ref{fig:graphic_overview} and Algorithm 1, both with and without active learning (\S \ref{sec:active}), to learn the optimal action. We model the expected utility $u(\bm{a}, \bm{x})$ as a GP with a square-exponential kernel, with separate correlation-length parameters for each input dimension. For the case with active learning, the neural amortized posterior is parameterized by a Mixture Density Network \cite{bishop1994mixture} with three Gaussian components, two hidden layers with $64$ units each, and $\mathrm{tanh}$ activation functions. Simulations are run in batches of $8$ and the GP hyper-parameters\footnote{With a square-exponential kernel, the complete set of hyper-parameters for the GP are the amplitude and correlation-lengths of the kernel, and the variance parameter $\sigma^2$.} and neural amortized posterior are re-trained (via gradient-desscent) after every new batch of simulations. In the active learning case, model parameters and actions for generating new simulations in each round are drawn from the neural amortized posterior and optimal action posterior respectively, otherwise they are drawn from their priors.

For a baseline to compare against, we solve the same decision problem using industry-standard posterior inference, Monte Carlo integration, and optimization techniques. To obtain the posterior, we use the same simulation-based inference set-up described above, i.e., sequentially fitting a Mixture Density Network to obtain the (amortized) posterior. The posterior expected utility integral (Equation \ref{eq:expected_utility}) for a given action $a$ is then estimated via Monte Carlo integration, averaging over $N=100$ posterior draws and simulated future outcomes\footnote{$N=100$ is chosen by trial-and-error so that the optimized action converges to within 1\% of the true optimal action.}. This estimator for the expected utility is then maximized with respect to action $a$ using differential evolution.
\subsection{Results}
The inferred optimal action as a function of the number of simulations, the learned utility surface and optimal action posterior, and the convergence of the amortized neural posterior, are shown in Fig.~\ref{fig:wh_results}. The number of simulations required for the optimal action to converge (to within $1\%$ of the true optimal action) is shown in Table \ref{tab:results}.

For SBD with active learning, we see that the optimal action converges after around $56$ simulations, and notably requires only half the number of simulations required to achieve a converged posterior for this problem. The traditional approach of posterior inference, Monte Carlo integration and optimization requires a factor of $\mathcal{O}(1000)$ more simulations to achieve the same accuracy on the optimal action. We find that without active learning, SBD requires roughly twice the simulations compared to the active learning case.
\section{Example II: Planning a deer cull}
\label{sec:lv}
\begin{figure*}[t!]
    \centering
    \includegraphics[width=0.9\textwidth]{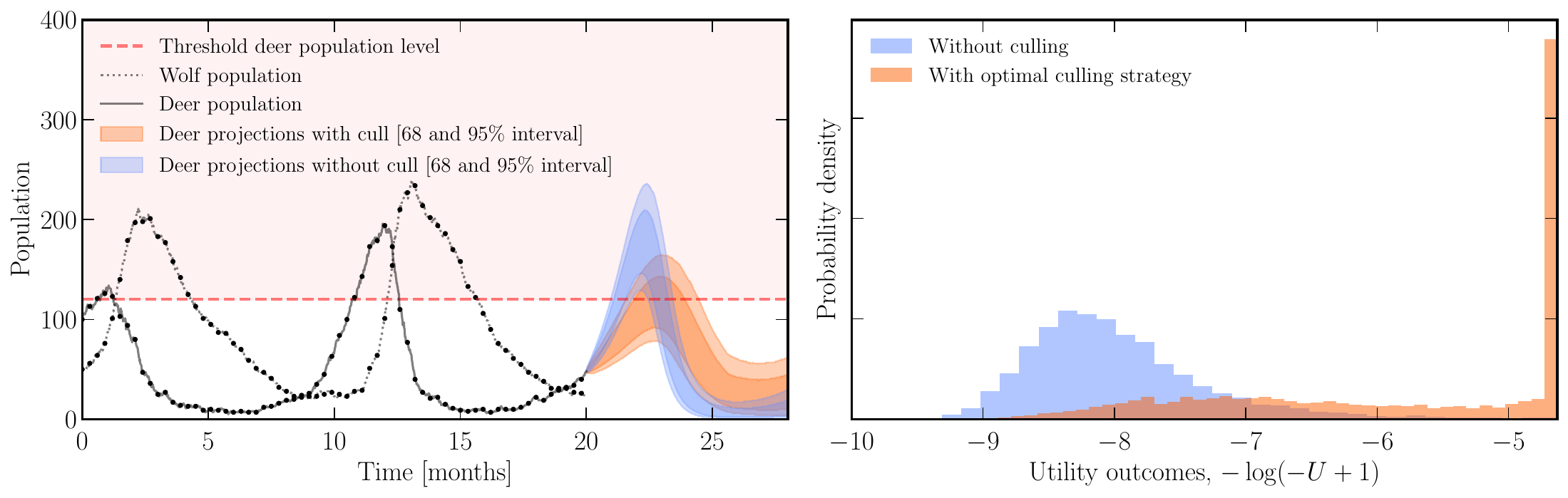}
    \caption{Left: simulated observations for the wolf (grey-dotted) and deer (grey) populations. Projections (68 and 95\% intervals) for the deer population are shown with (orange) and without (blue) culling (simulated assuming the true model parameters). Right: example utility distributions (simulated with the model parameters fixed to their true values) for scenarios with (orange) and without (blue) culling. The utility distribution in this case is highly non-Gaussian and heterogeneous across the action space.}
    \label{fig:lv_data}
\end{figure*}
Imagine you are the mayor of a small island nation and are faced with the following wildlife management problem. There are healthy fluctuating populations of deer and wolves living on the island, but also a number of farmers. When the deer population exceeds a certain level, $T$, natural food sources no longer sustain them and they start to eat the farmers crops at a cost of $C$ per excess deer (above threshold $T$) per day. In the case of a period of overpopulation, there are twenty people with hunting licenses on the island who could be called on to cull the deer, at a cost of $P$ per hunter per day.

Your advisors call a meeting to report that the deer population is rising rapidly and forecast to exceed $T$ within a few months (Figure~ \ref{fig:lv_data}), based on the data $\bm{x}$ (weekly observations of the deer and wolf populations for the past twenty months). You need to decide how many hunters to employ ($a_1$), and for how long $(a_2)$, to minimize the impact of a potential impending deer overpopulation, at minimum net cost over the next 12 months:
\begin{eqnarray}
U(\bm{y}, \bm{a}) = - a_1 P - C\int_0^{12} y(t) \Theta(y(t) - T)dt,
\end{eqnarray}
where $y(t)$ is the (outcome) future deer population as a function of time (whose distribution will depend implicitly on the actions $a_1$ and $a_2$), and $\Theta$ is the Heaviside function.

The populations of deer $D(t)$ and wolves $W(t)$ is assumed to follow the Lotka-Volterra model: wolves are born with rate $\alpha W(t)$, consume deer at a rate $\beta W(t) D(t)$, and die of natural causes at a rate $\delta W(t)$, while deer are born at a rate $\gamma D(t)$. When hunters $H$ are introduced into the system, they are assumed to kill deer at a rate of $r H D(t)$. The effectiveness of the hunters $r$ is assumed known, but all other model parameters $\bm{\theta} = (\alpha, \beta, \delta, \gamma)$ are unknown and need to be inferred from the observations.
\subsection{Experiments}
The observation vector consists of weekly observations of the deer and wolf populations for the past twenty months (Figure \ref{fig:lv_data}), simulated assuming $\alpha = 0.01$, $\beta = 0.5$, $\delta = 1$, $\gamma = 0.01$ and $r = 0.02$. For the utility, we take $P=1$, $C=50$ and $T=120$. We assume log-uniform priors on the Lotka-Volterra model parameters, and $P(a_1)=U(0, 20)$ and $P(a_2)=U(0,8)$ for the action priors.

We run the SBD algorithm described in Figure \ref{fig:graphic_overview} and Algorithm \ref{alg:algotirhm1} to learn the optimal action, both with and without active learning. Simulations are run in batches of $128$, and simulated observation vectors are compressed down to four summary statistics\footnote{Predator-prey time-series are first compressed down to nine summaries: the means, variances, lag-$1$ and lag-$2$ correlation coefficients of the deer and wolf population time-series, and the correlation coefficient between the deer and wolf populations. Those nine summaries are then further compressed to noisy parameter estimators for the four Lotka-Volterra model parameters, using a dense neural network with two hidden layers of $128$ units with leaky-ReLU activations, trained on simulations from the prior to minimize the mean-square error to the parameters.}. 
The neural amortized posterior is parameterized by a Mixture Density Network with six Gaussian components, and two hidden layers with $64$ units each, and $\mathrm{tanh}$ activation functions. For this problem, the utility is strictly negative, and is expected to be heavy-tailed (owing to the exponential behaviour of Lotka-Volterra like systems). We therefore model the expectation value of $(1-U)$ as a log-normal process, with a square-exponential kernel with separate length-scale parameters for each input variable. The log-normal process hyper-parameters, and neural amortized posterior, are re-fit (by gradient descent) after each new batch of simulations.

As a baseline to compare against, we again solve the same problem with the usual sequence of posterior inference, Monte Carlo integration and optimization. The posterior is obtained using the same simulation-based inference described above. The posterior expected utility integral (Equation \ref{eq:expected_utility}) for a given action $a$ is then estimated via Monte Carlo integration, averaging over $N=1000$ posterior draws and simulated future outcomes\footnote{$N=1000$ is chosen by trial-and-error so that the optimized action converges to within 10\% of the true optimal action.}. The expected utility is then maximized with respect to action $\mathbf{a}$ using differential evolution.
\begin{figure*}[t!]
    \centering
    \includegraphics[width=16cm]{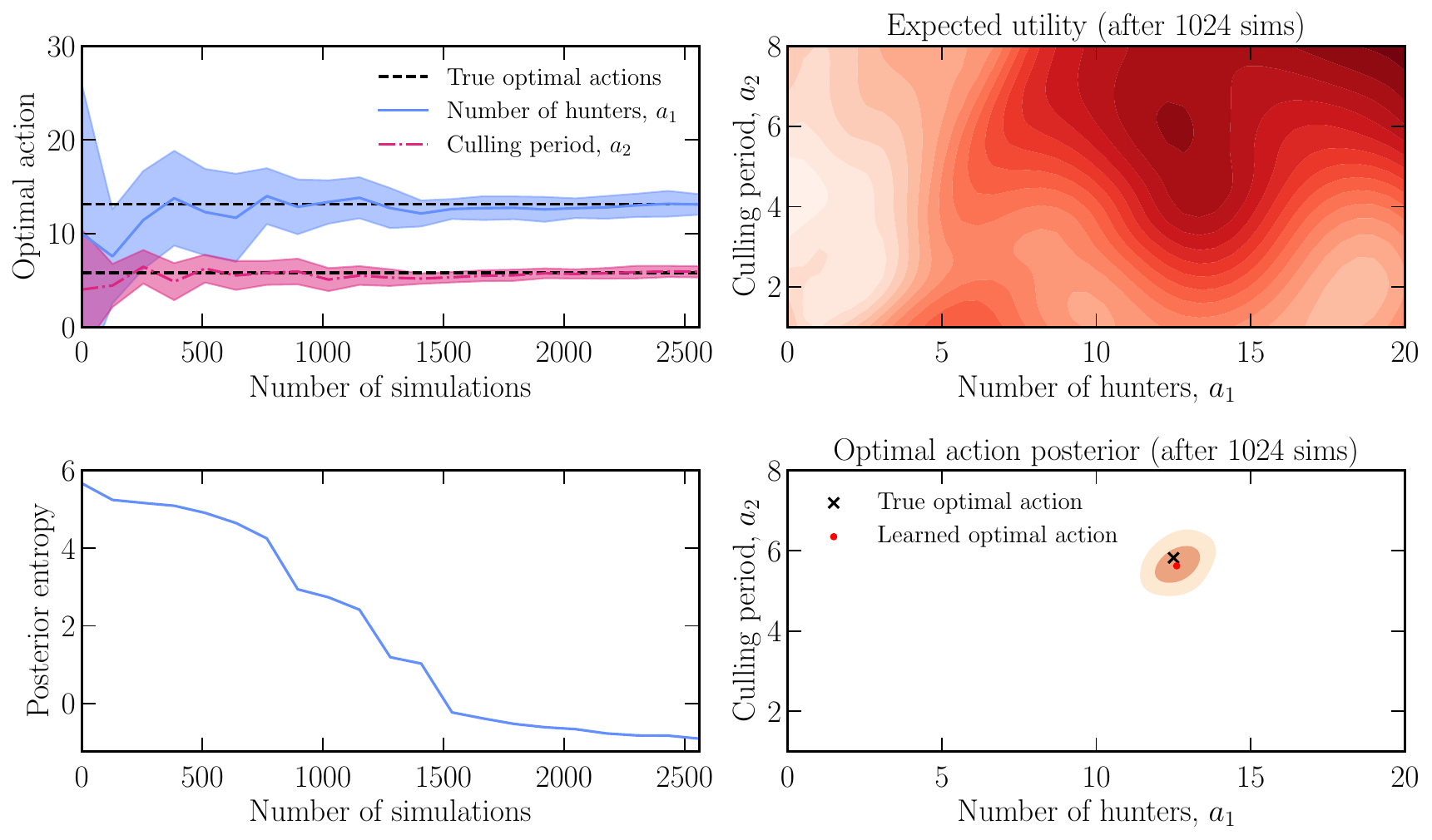}
    \caption{Top left: learned optimal action (with 68\% uncertainties) as a function of the number of simulations. Top right: learned expected utility surface after 1024 simulations. Bottom left: convergence of the amortized posterior (entropy) as a function of the number of simulations. Bottom right: obtained optimal action posterior (68 and 95\% intervals) after 1024 simulations, with the true optimal action shown for reference (black-cross).}
    \label{fig:lv_results}
\end{figure*}
\subsection{Results}
The inferred optimal action as a function of the number of simulations, the learned utility surface and optimal action posterior, and the convergence of the amortized neural posterior is shown in Fig.~\ref{fig:lv_results}. 

For SBD with active learning, the optimal action converges (to within $10\%$ of the true optimal action) after around $1024$ simulations -- roughly half the number required to obtain a converged neural posterior estimator for this problem. SBD required a factor of around $200$ fewer simulations compared to the baseline Monte Carlo based approach (for the same accuracy), and SBD with active learning required roughly a quarter of the simulations compared to the case without active learning (Table \ref{tab:results}).

Note that the relative uncertainties on the optimal number of hunters and culling period reflect how sensitive the expected utility is to those variables around the peak; in this case, the expected utility is less sensitive to the culling period, so the GP uncertainties converge more slowly for that parameter. We also note that both action parameters appear to have converged after around $1024$ simulations, even if the GP uncertainties remain non-negligible; the Gaussian errors in this case are likely to be conservative, owing to substantial heterogeneity and non-Gaussianity in the utility distribution across the action-space.
%
%

\section{Conclusions}
Leveraging recent advances in simulation-based inference (SBI), we have developed a new framework for computing optimal Bayesian decisions: simulation-based decision-making (SBD). We have shown that Bayesian decision making can be cast as a regression or conditional density-estimation task, learning a surrogate model for the expected utility surface (over the action- and data- spaces) from simulations, which can then be optimized with respect to the action to find optimal actions. Phrasing decision problems as regression or density-estimation tasks in this fashion circumvents having to either perform posterior inference, or compute (high-dimensional) integrals for the posterior predictive expected utility, explicitly; both of these steps are taken care of implicitly. Since no explicit likelihood calculations are required, the approach works equally well in the explicit and implicit likelihood regimes.

We have introduced two active learning schemes to reduce the overall simulation burden of computing optimal decisions. Training an amortized neural posterior estimator on-the-fly provides an adaptive proposal density for drawing model parameters for new simulations, therefore ensuring our efforts are concentrated in the neighbourhood of the observed data (rather than over the full prior-predictive support in data space). Secondly, by performing Bayesian inference of the utility surface itself, we have at any given stage a posterior density for the optimal action, given the training data obtained so far. This provides a natural proposal density for drawing which actions to simulate next, in order to zero in on the neighbourhood of the optimal action efficiently (i.e., Bayesian optimization).

Two important features of our approach bear discussing: first, beyond simply giving a point estimate of the optimal decision, our approach also quantifies the uncertainty in the optimal decision given the set of simulations provided so far. 

Second, and more generally, our approach does not stop at inferring optimal decisions, but also maps out the expected utility (and uncertainty on it) over the space of possible decisions. Therefore, our approach allows one to assess what is known about the expected utility. If the expected utility has a broad maximum, or secondary maxima, this could show that other decisions are nearly as good as the optimal decision. In a real world situation, human inspection of the utility surface may allow for adapting to unforeseen factors, or finding alternative decisions that take into account factors not explicitly modelled in the utility function in the first instance. This represents a significant advantage over black-box approaches (such as reinforcement learning) which just provide a single optimal action estimate.

The resulting framework for mapping out the expected utility function and computing optimal Bayesian decisions is extremely simulation efficient, typically requiring fewer model calls than would be required to perform the posterior inference step alone, and a factor of $100-1000$ fewer than would be required by the standard approach of performing optimization with respect to a Monte-Carlo estimate of the posterior predictive expected utility. In both the examples discussed in \S~\ref{sec:wh}-\ref{sec:lv}, the optimal action converges significantly faster than the associated posterior inference task (Figs.~\ref{fig:wh_results} and \ref{fig:lv_results}), requiring roughly half the simulation budget in those cases. This efficiency enables new capabilities for performing Bayesian decision making, particularly in the regime where likelihoods are intractable and simulations expensive.

One remaining challenge with our approach is that, once a surrogate model for the utility surface has been obtained (as a function of action and observations), an optimization step is still required to obtain the optimal action. For problems where the expected utility is convex with respect to the action, this final optimization step is cheap and easy with gradient-based optimization methods. However, high-dimensional non-convex problems with many locally-optimal actions may provide more of a challenge. Even in these more challenging cases, our approach of learning a computationally efficient surrogate model for the expected utility over the action and observation spaces would still provide significant advantages over performing the equivalent optimization problem using noisy, point-wise Monte-Carlo estimates for the expected utility.

\bibliography{main}
\bibliographystyle{icml2021}

\end{document}